\PassOptionsToPackage{expansion=false,protrusion=false}{microtype}
\documentclass{ceurart}
\usepackage{listings}
\newif\ifshowcolors
\ifshowcolors
  \definecolor{revcolor}{rgb}{0.8,0,0}
  \definecolor{origcolor}{rgb}{0,0.5,0}
  \definecolor{thesiscolor}{rgb}{0,0,0.85}
  \definecolor{humancolor}{rgb}{0.55,0,0.75}
\else
  \definecolor{revcolor}{rgb}{0,0,0}
  \definecolor{origcolor}{rgb}{0,0,0}
  \definecolor{thesiscolor}{rgb}{0,0,0}
  \definecolor{humancolor}{rgb}{0,0,0}
\fi
\newcommand{\revtxt}[1]{{\color{revcolor}#1}}
\newcommand{\origtxt}[1]{{\color{origcolor}#1}}
\newcommand{\thesistxt}[1]{{\color{thesiscolor}#1}}
\newcommand{\humantxt}[1]{{\color{humancolor}#1}}

\begin{document}

\copyrightyear{2026}
\copyrightclause{Copyright for this paper by its authors.
  Use permitted under Creative Commons License Attribution 4.0
  International (CC BY 4.0).}

\conference{13th Italian Workshop on Artificial Intelligence and Robotics (AIRO 2026)}

\title{iAm.md: Robot Skill Self-Assessment through Agentic Introspection for Unknown Open-Vocabulary Domains}

%% All author ORCIDs filled
\author[1]{Vincenzo Guarino}[%
orcid=0009-0006-6794-6767,
email=vincenzo.guarino@uniroma1.it,
]
\cormark[1]
\author[1]{Emanuele Musumeci}[%
orcid=0009-0004-2359-5032,
email=musumeci@diag.uniroma1.it,
]
\author[1]{Vincenzo Suriani}[%
orcid=0000-0003-1199-8358,
email=suriani@diag.uniroma1.it,
]
\author[1]{Daniele Nardi}[%
orcid=0000-0001-6606-200X,
email=nardi@diag.uniroma1.it,
]
\address[1]{Department of Computer, Control and Management Engineering ``Antonio Ruberti'', Sapienza University of Rome, Via Ariosto 25, 00185 Rome, Italy}

\cortext[1]{Corresponding author.}

\begin{abstract}
Agentic AI based on Large Language Model generalization capabilities offers a wide range of potential applications, including planning for embodied tasks. For example, embodied agents based on Foundation models can generate plausible plans in autonomous robotics scenarios. Due to limited context windows or hallucinatory phenomena in the next-token prediction formulation, behaviors may be generated without establishing whether the deployed robot and the observed environment actually support the requested operation, in what we call a "grounding failure". Thanks to the recent improvements in reasoning capabilities of foundation models, autonomous robot behavior generation problem can be formulated as a code generation problem. We present \texttt{iAm.md}, a Markdown standard and generation framework, that allows anchoring this process in complementary forms of deployment evidence. Through open-vocabulary semantic mapping, we combine local \humantxt{vision-language} detections and object segmentation and refer them to persistent object records in this intermediate standardized representation, allowing agentic introspection. We then study this new technique on a simulated TIAGo, on navigation-and-manipulation tasks, showing how this standardized representation jointly supports skill self-assessment and executable task generalization.
\end{abstract}


\maketitle

\section{Introduction}
\label{sec:intro}

Translating a Natural Language (NL) task request into physical robot behavior typically requires a closed representation of the task and operating environment: neuro-symbolic planning makes plan validity conditional on the quality of a provided abstraction of the world state and actions~\cite{fikes1971strips,mcdermott1998pddl,garrett2021tamp}. As robots move toward open environments, prompt engineering is an increasingly important step of plan and executable code generation~\cite{brown2020gpt3,chen2021codex}, but this capacity creates a grounding problem: a language model may complete an underspecified request with synthetic knowledge even when the necessary evidence has not been observed~\cite{ji2023hallucination}, and generated action sequences can violate the preconditions of the actions they invoke~\cite{valmeekam2023planbench}. Once connected to robot tools, an unsupported assumption becomes a physical action, which can lead to obvious safety and ethical issues.

In this way, a generated program may refer to an unobserved object, or it may call for an operation that the deployed platform does not support. Scene grounding alone does not resolve the second problem, since a plan reasonable for one robot may fail on another. Robot self-assessment frames the missing judgment as the ability to "predict, estimate, or measure how well it can perform a task in a given context and environment"~\cite{norton2022metrics}. We investigate to what extent an embodied foundation model agent operating in an unknown open-vocabulary domain can use persistent scene memory and a structured robot self-description to perform \emph{skill self-assessment} and produce executable plans. Here \emph{agentic introspection} means the agent is capable of inspecting the representations available inside its own planning process before committing to generated code. Our contributions are: (i) an open-vocabulary semantic memory that uses local \humantxt{vision-language} inference and segmentation to provide persistent object identities and locations; (ii) \texttt{iAm.md}\footnote{Project page: \url{https://yurimachine.github.io/iAm.md/}.}, a structured Markdown standard and generation framework representing specific robot information, resolved interfaces, and explicit unknowns; and (iii) an introspective embodied agent that combines semantic-map queries, introspection, task code generation, and robot execution for NL requests.

\section{Related Work}
\label{sec:related}

Hierarchical 3D scene graphs (Kimera~\cite{rosinol2021kimera}, Hydra~\cite{hughes2022hydra}) and dense \humantxt{vision-language} feature maps (VLMaps~\cite{huang2023vlmaps}, ConceptFusion~\cite{jatavallabhula2023conceptfusion}) accumulate persistent spatial structure but do not maintain robot-specific execution evidence. Language-conditioned code-generation methods (SayCan~\cite{ahn2022saycan}, Code as Policies~\cite{liang2023cap}, ProgPrompt~\cite{singh2023progprompt}) compose developer-provided APIs but assume feasible instructions and pre-authored platform documentation; feasibility-assessment methods (PlanBench~\cite{valmeekam2023planbench}, CodeBotler~\cite{hu2024codebotler}, KnowNo~\cite{ren2023knowno}) require a complete action model rarely available for a configured robot. Robot knowledge frameworks~\cite{tenorth2017knowrob,bessler2021soma,silva2025rosa} relate tasks to capabilities through data entered before an episode. Closest to our work, EMPOWER-ADF~\cite{suriani2026empower} grounds task entities against a NL robot description but selects from a fixed action vocabulary without physical execution. We instead derive the robot description from the deployment itself and use it, with live interface evidence, to \emph{write} task-specific code.

\section{\revtxt{Methodology}}
\label{sec:approach}

\begin{figure}[t]
\centering
\includegraphics[width=0.78\linewidth]{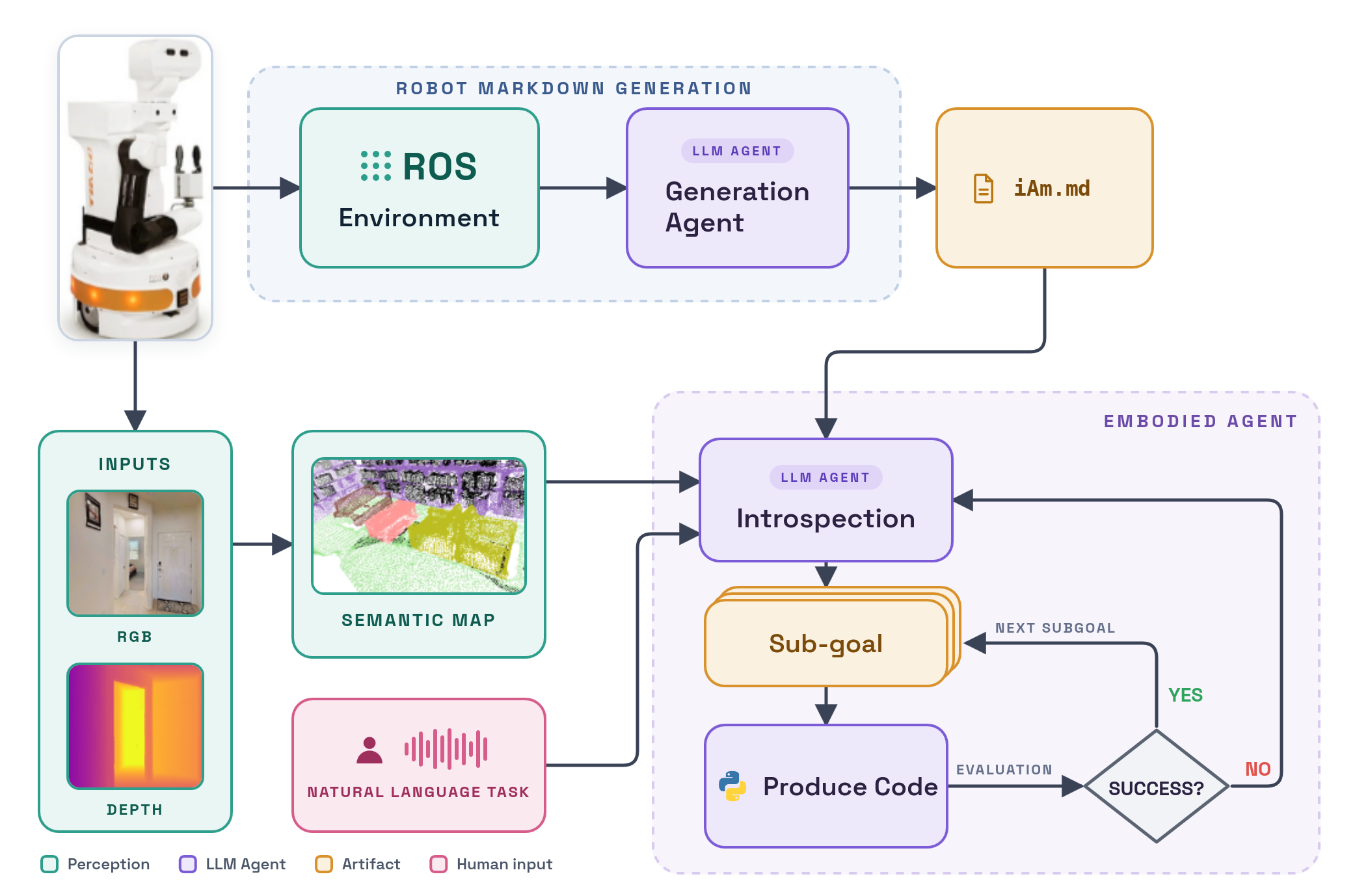}
\caption{\textcolor{revcolor}{System architecture of the robot-description generation and embodied-agent execution stages. The generation agent extracts resolved robot evidence from the ROS~2 environment and records it in \texttt{iAm.md}. During task execution, the embodied agent combines this description with the semantic map and NL request, produces code for successive subgoals, and uses execution outcomes for introspective revision.}}
\label{fig:arch}
\end{figure}

An embodied code generation system places a tool-calling foundation-model agent between a task request and the  deployed robot. Persistent robot facts enter the agent context through \texttt{iAm.md}, whereas the current pose and camera observation describe the state in which the request must be executed. One general command tool connects this context to the live ROS~2 environment~\cite{macenski2022ros2}, allowing the agent to observe the execution state, write executable programs, and run them in the deployment that supplied the supporting evidence. The semantic map supplies object identities and spatial support through ROS~2 topics and a serialized snapshot, so generated programs resolve task targets without placing transient observations in the persistent robot description. \revtxt{Figure~\ref{fig:arch} summarizes these evidence paths and the execution loop that connects the generated program to the deployed robot.}

\paragraph{Open-vocabulary semantic memory.} A locally served \humantxt{vision-language} model (Qwen3.6-35B via llama.cpp) proposes whole-object detections with open-vocabulary labels and bounding boxes; after non-maximum suppression, SAM~2~\cite{ravi2024sam2} produces a binary mask for each surviving proposal. Each validated mask is reprojected through depth into a map-frame point cloud, cleaned by morphological closing, foreground filtering, and DBSCAN clustering~\cite{ester1996dbscan}, and voxelized into a sparse \texttt{VoxelBlockGrid}~\cite{zhou2018open3d} whose active voxels carry object identifiers. \revtxt{Association is decided by a voxel-overlap score between the new observation $A$ and each stored support $B$,}
{\color{revcolor}
\begin{equation}
s(A,B) = \max\!\left(\frac{|A \cap B|}{|A|}, \frac{|A \cap B|}{|B|}\right),
\end{equation}}
\revtxt{so an observation below threshold for every candidate becomes a new object, a single match updates the stored record, and multiple matches trigger a merge.} Stored voxels yield centroids, bounds, and oriented bounding boxes exported as a JSON snapshot and RViz markers that downstream programs query to resolve NL references, including entities outside the current camera view.

\paragraph{\texttt{iAm.md}: embodiment evidence.} \origtxt{\texttt{iAm.md} is a standard for describing the persistent embodiment of a configured robot deployment in plain text that a foundation-model agent can read in context. A conforming document follows a fixed schema from identity and embodiment summary, through the URDF-derived physical structure, actuation, and sensing, to the higher-level software functions exposed by the robot stack and the unknowns that remain after generation. Every block carries explicit cardinality, a value that cannot be defensibly established is recorded as \emph{unknown} in place, and software-function records separate the \emph{acceptance} criteria of a command interface from the physical \emph{outcome} of the commanded operation. An instance is generated once per deployment from the URDF, augmented with the live ROS~2 graph and configuration artifacts for otherwise-unobservable interfaces. The TIAGo~\cite{pages2016tiago} instance used here records its kinematic structure, navigation and gripper envelopes, controller bindings, and Navigation2~\cite{macenski2020marathon}/MoveIt function records. The \humantxt{acceptance/outcome} separation appears concretely in the gripper records: a successful gripper trajectory result establishes only that controller tolerances were met, and grasp retention is listed among the unknowns because the deployed configuration includes no sensor that could establish it.}

\paragraph{Agentic introspection and robot code execution.} \origtxt{Each NL request is one execution episode. The startup prompt defines the command-tool protocol and embeds the generated \texttt{iAm.md} between explicit delimiters. The agent inspects the ROS~2 graph and interface schemas through the same command tool it uses to act, so interface availability is established from the deployed system during the episode. Skill self-assessment runs inside the agent loop: the agent relates the request to observed objects, then checks the robot description and live interfaces needed to realize the \humantxt{operation/navigation} needs the object location and navigation interface, whereas manipulation additionally requires the manipulator and gripper contracts. When evidence supports the operation the agent proceeds to observation, program construction, or a physical command; otherwise a structured outcome records a blocked result. Introspection continues during execution, because observed behavior can expose an unsupported assumption and prompt a revision of the program.}

\section{Experimental Results}
\label{sec:results}

\subsection{Semantic-Map Evaluation}

\origtxt{Twelve objects in a simulated scene were independently annotated as labelled spheres in the map frame through an interactive RViz interface. An estimated record is compatible with an annotation when its centroid lies inside the sphere and its label matches; a maximum one-to-one bipartite matching over compatible pairs yields object-level \emph{Found} (fraction of annotations matched) and \emph{Correct} (fraction of estimates matched) scores, following Hydra~\cite{hughes2022hydra}. The matching selected 12 pairs from 12 annotations and 13 estimated records: every annotated object was recovered (Found $=100\%$), and the single unmatched record was a duplicate estimate (Correct $=92.31\%$), localizing the observed error to duplicate consolidation during map updates and merges.}

\subsection{Agent-Produced Tool Evaluation}

The agent evaluation isolates the contribution of \texttt{iAm.md} by varying only the persistent robot knowledge available during tool construction. \thesistxt{Both conditions run the same embodied coding agent, a DeepAgents harness driving \texttt{gpt-5.6-sol} at medium reasoning effort, on a simulated single-arm TIAGo, and both receive the same fixed request together with a distilled NL specification of the navigation-and-manipulation tool to be built. The specification was distilled from a reference implementation developed with Claude Fable~5 on the same platform, with every deployment fact removed, so the agent must recover the robot connection itself, and \texttt{iAm.md} is placed only in the documented workspace.} A phase ends when the submitted tool passes independent verification (a simulator reset, unattended execution, and a static audit rejecting tools that do not use robot depth data), or when the agent reaches a 110-minute budget after at most four attempts.

\paragraph{Phase 1: construction.} \origtxt{For \emph{``grasp a drink can and place it on a sofa''}, the documented agent produced a tool that passed verification after a simulator reset in 50.4 minutes, using 103 model calls and 8.50 million input tokens. The agent without \texttt{iAm.md} reached the 110-minute budget (186 model calls, 20.32 million input tokens) without a verified grasp or a completed task (Table~1A). The gap appeared at the earliest milestones: the documented agent wrote a first tool version after 4.8 minutes and reached a verified grasp approach after 7.8 minutes, against 8.2 and 26.4 minutes without the description, and it first verified a grasp after 25.7 minutes while the other agent obtained none within budget (4/10 versus 0/19 successful grasp attempts). Lacking the recorded embodiment, the undocumented agent calibrated the gripper through execution and incurred 13 initial inverse-kinematics failures, recovering constraints that \texttt{iAm.md} supplies directly; it also made 23 live-discovery calls, against 14 discovery calls and five \texttt{iAm.md} reads for the documented agent.}

\paragraph{Phase 2: reuse.} \origtxt{Retaining the documented agent's state and tool, the request changed to \emph{move a cube to the opposite side of a table}, a different object geometry and placement relation. The agent raised the torso, verified clearance, and executed the verified joint configuration, avoiding an elbow-wrapping IK switch during retreat. To represent \emph{the other side}, it extended the tool interface with a map-frame approach direction rather than a cube-specific behavior. The plan passed its first verification after a simulator reset. Reuse required only three additional discovery calls and no further \texttt{iAm.md} reads (Table~1B); the changed geometry nonetheless propagated into grasp planning and placement, requiring 49 tool edits and 34 reported step failures. Retaining the command vocabulary while extending the generic kinematic and directional mechanisms constitutes few-shot generalization.}

\begin{table}[t]
\centering
\caption{(A) Phase 1 construction with and without \texttt{iAm.md}; the condition without the document reports the effort before the 110-minute budget was reached. \textcolor{revcolor}{(B) The documented agent across construction (P1) and reuse (P2).}}
\label{tab:phase1}
\begin{minipage}[t]{0.48\linewidth}
\centering
\textbf{Table 1A}\par\vspace{2pt}
\scriptsize
\setlength{\tabcolsep}{3pt}
\color{thesiscolor}
\begin{tabular}{lccc}
\toprule
Metric & With & Without & Ratio \\
\midrule
Verified task completion & Yes & No (budget) & \humantxt{-} \\
Time to verified compl. (min) & 50.4 & $>$109.9 & $>$2.2$\times$ \\
Model calls & 103 & 186 & 1.8$\times$ \\
Input tokens (million) & 8.50 & 20.32 & 2.4$\times$ \\
Tool edits & 33 & 47 & 1.4$\times$ \\
Reported step failures & 22 & 48 & 2.2$\times$ \\
Successful grasp attempts & 4/10 & 0/19 & \humantxt{-} \\
\bottomrule
\end{tabular}
\end{minipage}\hfill
\begin{minipage}[t]{0.48\linewidth}
\centering
\textbf{\textcolor{revcolor}{Table 1B}}\par\vspace{2pt}
\scriptsize
\setlength{\tabcolsep}{3pt}
\color{thesiscolor}
\begin{tabular}{lcc}
\toprule
Metric & P1 & P2 \\
\midrule
Verified completion & Yes & Yes \\
Time to verified (min) & 50.4 & 94.5 \\
Model calls & 103 & 150 \\
Input tokens (M) & 8.50 & 16.90 \\
Tool edits & 33 & 49 \\
Grasp attempts & 4/10 & 8/13 \\
Discovery calls & 14 & 3 \\
\texttt{iAm.md} reads & 5 & 0 \\
\bottomrule
\end{tabular}
\end{minipage}

\vspace{0.8em}
\centering
\includegraphics[width=0.32\linewidth]{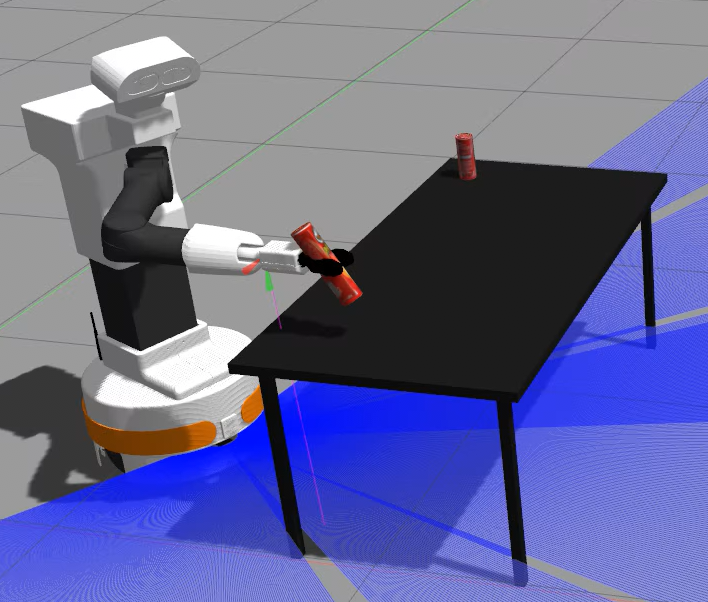}\hspace{1.5em}%
\includegraphics[width=0.32\linewidth]{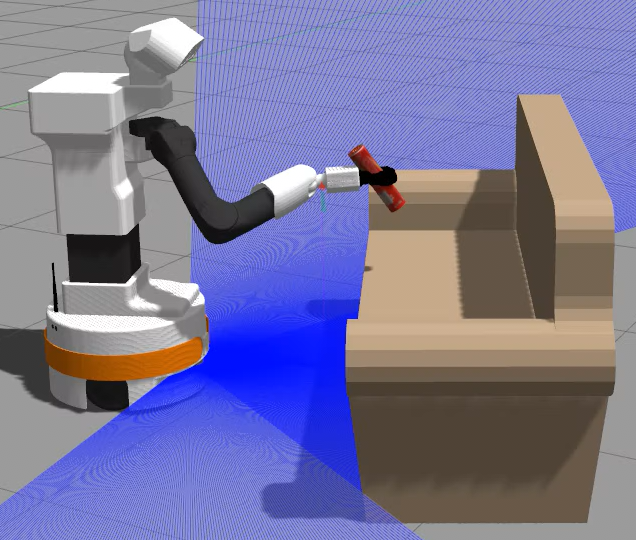}
\ExplSyntaxOn\dim_set:Nn \l_fig_width_dim { \linewidth }\ExplSyntaxOff
\captionof{figure}{\textcolor{revcolor}{\small Simulated TIAGo executing the agent-produced tool: picking up the can (left) and placing it on the sofa (right).}}
\label{fig:exec}
\end{table}

\subsection{Discussion}

\origtxt{The two evaluations identify complementary roles for persistent and live evidence during skill self-assessment. The semantic map supplies the object referents and spatial estimates needed to interpret a request; \texttt{iAm.md} relates those estimates to the configured robot; and the resulting assessment is expressed through executable checks that change when execution exposes an unsupported assumption. Because checks derived from failures remained part of the tool and were reused when the request changed, self-assessment emerged as a feedback loop connecting task interpretation, code generation, execution evidence, and revision.}

\section{Conclusions and Future Work}
\label{sec:conclusions}

\revtxt{We showed that an embodied foundation-model agent can combine persistent scene memory, a robot self-description, and live ROS~2 observations to assess and realize NL tasks in an open-vocabulary domain. The semantic map recovered all twelve annotated objects with one duplicate record. In a controlled comparison, access to \texttt{iAm.md} was associated with lower construction cost and verified completion: the documented agent produced a verified tool in 50.4 minutes, versus the 110-minute budget without a verified grasp for the undocumented agent, expending roughly twice the calls and tokens. Retaining that tool, the agent extended it to a second task with different object geometry without rereading \texttt{iAm.md}\humantxt{-}evidence of few-shot generalization, from a single simulated run. Future work will evaluate \texttt{iAm.md} generation across embodiments and add an independent safety boundary validating trajectories before actuation on physical robots.}

%\begin{acknowledgments}
%\origtxt{This work is based on the first author's Master's thesis at Sapienza University of Rome.}
%\end{acknowledgments}

\begin{aideclaration}
During the preparation of this work, the authors used a large language model (LLM)-based assistant to proofread and optimize the phrasing of parts of this manuscript, including language editing, sentence-level optimization, and ambiguity resolution. After using this tool, the authors reviewed and edited the content as needed and take full responsibility for the content of the publication.
\end{aideclaration}

\bibliography{refs}

\end{document}